\documentclass[11pt, a4paper]{article}
\usepackage{arxiv}
\usepackage[utf8]{inputenc}
\usepackage[T1]{fontenc}
\usepackage[sort&compress,numbers]{natbib}
\usepackage{xcolor}
\definecolor{blendedblue}{rgb}{0.2, 0.2, 0.6}
\usepackage{array, multirow, graphicx}
\usepackage{enumerate}
\usepackage[bookmarks   = true,
            citecolor   = blendedblue,
            colorlinks  = true,
            linkcolor   = blendedblue,
            urlcolor    = blendedblue,
            citecolor   = blendedblue,
            linktocpage = false,
            hyperindex  = true]{hyperref}
\usepackage{booktabs}
\usepackage{url}
\usepackage{doi}
\usepackage{orcidlink}
\usepackage{fontawesome}
\usepackage{lmodern}
\usepackage{booktabs}
\usepackage{enumitem}
\usepackage{rotating}
\usepackage{diagbox}
\usepackage{xltabular}
\usepackage{threeparttablex}
\usepackage{afterpage}

\def\BibTeX{{\rm B\kern-.05em{\sc i\kern-.025em b}\kern-.08em
    T\kern-.1667em\lower.7ex\hbox{E}\kern-.125emX}}
\newcommand*\ec[1][0.7ex]{\tikz\draw (0,0) circle (#1);} 
\newcommand*\hc[1][0.7ex]{%
  \begin{tikzpicture}
  \draw[fill] (0,0)-- (90:#1) arc (90:270:#1) -- cycle ;
  \draw (0,0) circle (#1);
  \end{tikzpicture}}
\newcommand*\fc[1][0.7ex]{\tikz\fill (0,0) circle (#1);}

\def\BibTeX{{\rm B\kern-.05em{\sc i\kern-.025em b}\kern-.08em
    T\kern-.1667em\lower.7ex\hbox{E}\kern-.125emX}}

\usepackage[labelsep=period]{caption}

\renewcommand{\headeright}{}

\usepackage{parskip}
\makeatletter 
\renewcommand\@biblabel[1]{#1.} 
\makeatother

\title{Artificial Intelligence in Industry 4.0: A Review of Integration Challenges for Industrial Systems}

\renewcommand{\shorttitle}{AI in Industry 4.0: A Review of Integration Challenges for Industrial Systems}
\date{\vspace{1cm}} 

\author{
	Alexander Windmann\orcidlink{0000-0002-6522-4262} \\
 	Institute of Artificial Intelligence\\
        Helmut Schmidt University\\
	Hamburg, Germany\\
	\texttt{alexander.windmann@hsu-hh.de} \\
	\And
	Philipp Wittenberg\orcidlink{0000-0001-7151-8243} \\
 	Dep. of Mathematics and Statistics\\
        Helmut Schmidt University\\
	Hamburg, Germany\\
	\texttt{pwitten@hsu-hh.de} \\
	\And
	Marvin Schieseck\orcidlink{0009-0000-7941-5900} \\
        Institute of Automation Technology\\
        Helmut Schmidt University\\
	Hamburg, Germany\\
	\texttt{marvin.schieseck@hsu-hh.de}\\
	\And 
	Oliver Niggemann\orcidlink{0000-0001-8747-3596} \\
 	Institute of Artificial Intelligence\\
        Helmut Schmidt University\\
	Hamburg, Germany\\
	\texttt{oliver.niggemann@hsu-hh.de} \\
}

\begin{document}	
\maketitle

\begin{abstract}
In Industry 4.0, Cyber-Physical Systems (CPS) generate vast data sets that can be leveraged by Artificial Intelligence (AI) for applications including predictive maintenance and production planning. However, despite the demonstrated potential of AI, its widespread adoption in sectors like manufacturing remains limited. Our comprehensive review of recent literature, including standards and reports, pinpoints key challenges: system integration, data-related issues, managing workforce-related concerns and ensuring trustworthy AI. A quantitative analysis highlights particular challenges and topics that are important for practitioners but still need to be sufficiently investigated by academics.
The paper briefly discusses existing solutions to these challenges and proposes avenues for future research. We hope that this survey serves as a resource for practitioners evaluating the cost-benefit implications of AI in CPS and for researchers aiming to address these urgent challenges.
\end{abstract}


\vspace{.6cm}
\begin{center}
\fbox{
  \begin{minipage}{\textwidth-.9cm}
    \tiny{
      © 2024 IEEE.  Personal use of this material is permitted.  Permission from IEEE must be obtained for all other uses, in any current or future media, including reprinting/republishing this material for advertising or promotional purposes, creating new collective works, for resale or redistribution to servers or lists, or reuse of any copyrighted component of this work in other works. DOI: \href{https://www.doi.org/10.1109/INDIN58382.2024.10774364}{10.1109/INDIN58382.2024.10774364}}
  \end{minipage}
}    
\end{center}

\section{Introduction}\label{sec:intro}

In recent years, Artificial Intelligence (AI) has achieved remarkable success in diverse fields such as natural language processing \cite{Vaswani.etal_2017}, computer vision \cite{Dosovitskiy.etal_2021}, and bioinformatics \cite{Jumper.etal_2021}. These advancements highlight the potential of AI in processing complex and high-dimensional data, making it well-suited for enhancing automation in industrial Cyber-Physical Systems (CPS), thus accelerating the transformation toward Industry 4.0.
Despite this potential, the integration of AI into sectors like manufacturing remains limited \cite{Merkel-Kiss.VanGarrel_2023, Maschler.etal_2022}. According to a 2023 survey by the Boston Consulting Group, 89\% of manufacturers regard AI as essential, yet only 68\% have initiated implementing AI use cases. Of these, merely 16\% have met their AI-related targets \cite{WEF.BCG_2023}. This can be partially explained by the inherent challenges of implementing AI projects, such as difficulties in building infrastructure and a lack of expertise \cite{Paleyes.etal_2022}. Additionally, the application of AI in Industry 4.0 presents compounded challenges, such as stringent safety requirements of technical systems that are integrated with 'black-box' AI models \cite{Perez-Cerrolaza.etal_2024} and a lack of widely available, relevant and high-quality data \cite{Maschler.etal_2022}.

Understanding the challenges of integrating AI in industrial systems is crucial for practitioners to conduct thorough cost-benefit analyses and for researchers to prioritize the most urgent issues. While there is extensive research on challenges of deploying AI in general \cite{Sculley.etal_2015, Lwakatare.etal_2019, Paleyes.etal_2022}, industrial applications present unique challenges not covered by these works. Initial foundational studies addressed CPS challenges \cite{Lee_2008, Gunes.etal_2014, Broy.Schmidt_2014, Vogel-Heuser.etal_2014, Sisinni.etal_2018} but overlooked the complexities introduced by AI technologies. Quantitative surveys of practitioners offer a broad view of emerging issues yet fail to identify which challenges have been thoroughly explored in scholarly literature \cite{Machado.etal_2019, Merkel-Kiss.VanGarrel_2023, Ishikawa.Yoshioka_2019}. Furthermore, much of the current research focuses on addressing specific subsets of these challenges \cite{He.etal_2021, Lavin.etal_2022, Maschler.etal_2022}, without providing a holistic perspective that could effectively guide both practitioners and researchers in designing a comprehensive solution approach. Our study aligns with recent surveys on the integration of AI in Industry 4.0. However, many current studies lack a focused examination of the underlying challenges, often discussing related topics such as safety engineering \cite{Pereira.Thomas_2020,Perez-Cerrolaza.etal_2024}, Big Data \cite{Jagatheesaperumal.etal_2022}, cyber-threats \cite{Becue.etal_2021}, MLOps \cite{Faubel.etal_2023}, and AI-human workforce collaboration \cite{Gabsi_2024}. While some survey papers provide overviews of AI applications in Industry 4.0 and touch on certain challenges \cite{Xu.etal_2022, Jan.etal_2023, Chen.etal_2023}, they do not comprehensively map these challenges to the underlying literature, including essential standards and regulatory body reports. Moreover, there is a notable absence of quantitative analysis on research interest in specific challenges. This gap underscores the need for a detailed examination of the challenges associated with integrating AI in Industry 4.0.

\begin{figure}[!htb]
\centering
\includegraphics[scale=1]{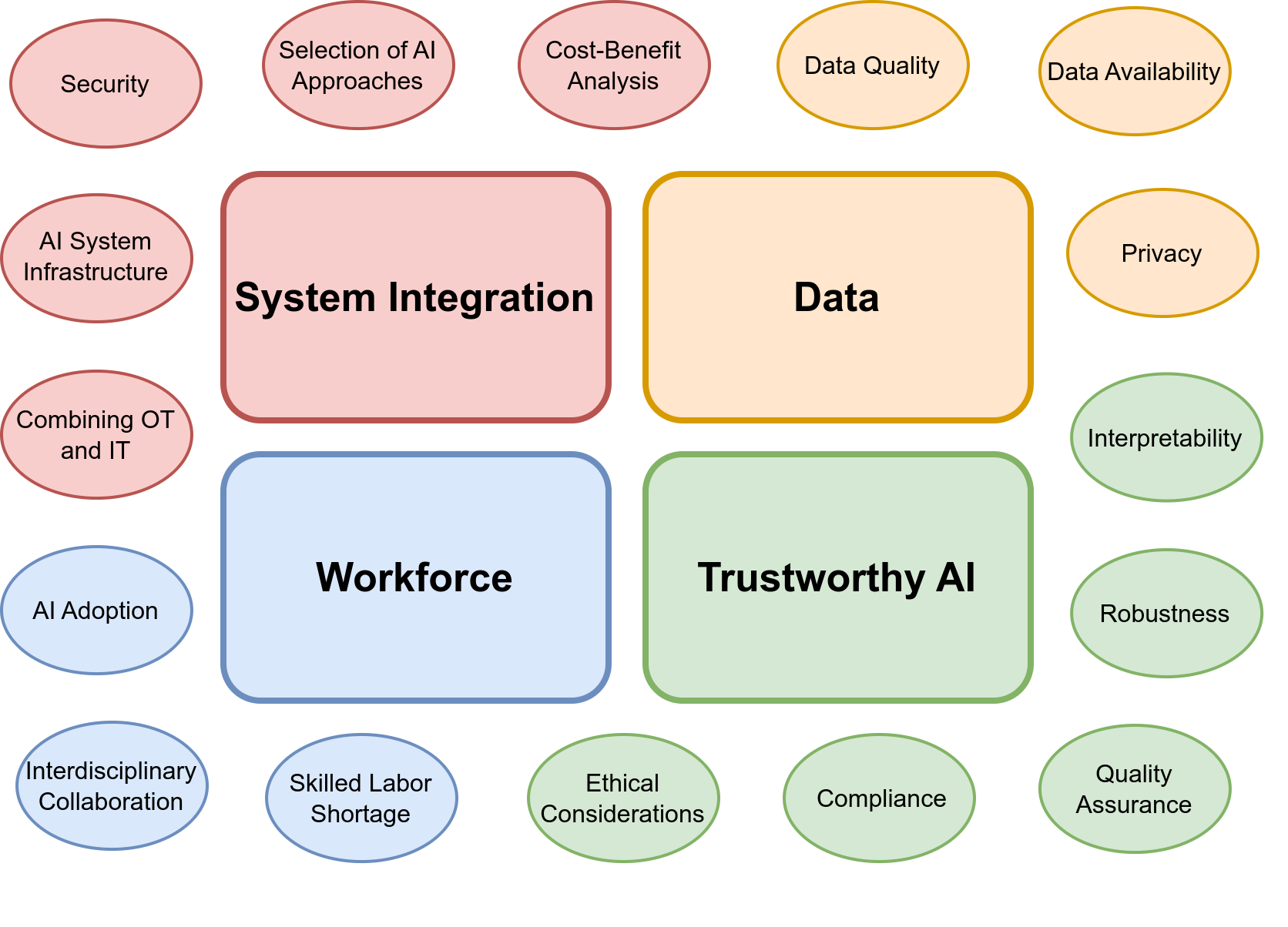}
\caption{Identified challenges of integrating AI within industrial systems.}
\label{fig:challenges}
\end{figure}

Our review addresses this need by systematically mapping the identified challenges to the relevant literature and quantitatively analyzing research trends. We suggest that this review could serve as a useful resource for understanding AI integration in industrial systems, providing insights beneficial to both industry practitioners and the academic community. The challenges of integrating AI in industrial systems identified by this survey are depicted in Figure \ref{fig:challenges}.

Our main contributions are:
\begin{itemize}
    \item We review the literature, including standards and reports, to comprehensively identify and categorize challenges related to AI integration in industrial systems, linking each challenge to the sources for a thorough overview.
    \item We conduct a frequency analysis of these challenges across sources to determine current research focus.
    \item We highlight gaps where identified challenges are recognized as pressing by practitioners but seldom addressed by the current literature.
\end{itemize}

The rest of the paper is structured as follows: 
Section \ref{sec:lit_search_methology} gives an overview of how relevant literature was identified. In Section \ref{sec:challenges}, we highlight the specific challenges that affect AI projects in industrial applications and give a comprehensive overview of the sources addressing these challenges. Section \ref{sec:discussion} details the quantitative results of the literature review, thus highlighting both common and underrepresented challenges. Finally, Section \ref{sec:conclusion} summarizes this article and proposes avenues for future research.

\section{Literature Search Methodology}\label{sec:lit_search_methology}
\begin{figure}[!htb]
\centering
\includegraphics[scale=.8]{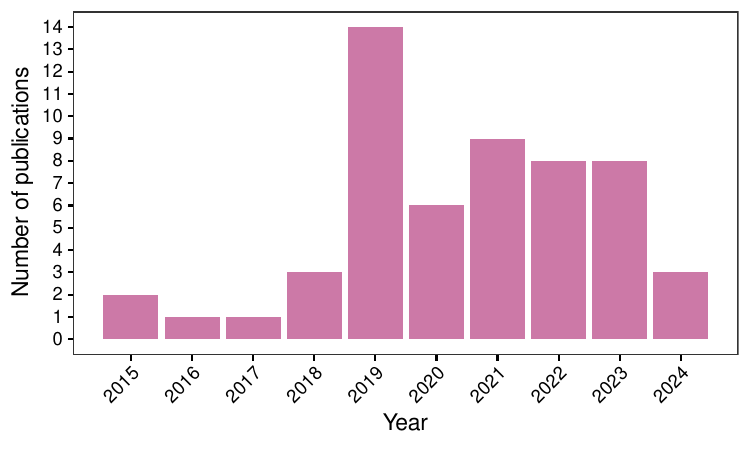}
\caption{Number of publications per year included in this study (sample size $n=55$). The full list of references can be found in Table \ref{tab:challengesA}.}
\label{fig:num_of_papers}
\end{figure}

The primary objective of this article is to provide a comprehensive overview of the current challenges in applying AI within industrial settings. To accomplish this, we systematically reviewed a diverse array of sources, including scientific articles, surveys, reports, white papers, and industry standards, with a focus on publications from the last decade. This timeframe was chosen to highlight recent challenges and ensure relevance to contemporary industrial applications. The distribution of the publications included in the analysis over time can be found in Figure \ref{fig:num_of_papers}.

Our literature review was conducted via academic search platforms including Google Scholar, Semantic Scholar, and Zeta Alpha. The keywords used were "challenges," "artificial intelligence," "machine learning," "engineering," "Industry 4.0," and "cyber-physical systems." We complemented this search by exploring relevant standards on the websites of the International Organization for Standardization (ISO), the International Electrotechnical Commission (IEC) and the German Institute for Standardization (DIN). We also reviewed cross-references in key papers, expanding our collection to include critical works that might otherwise have been missed. Our team's expertise, informed by ongoing AI projects, helped refine our search to include studies that provide deep insights into specific challenges.

We employed a strict screening and filtering process to ensure relevance. We only included sources that address challenges related to the deployment of AI, either broadly or within industrial systems. Sources that primarily discuss general challenges of CPS, such as hardware limitations and real-time data processing demands, were not considered for this study. A total of 55 sources passed through this filtering process and are discussed in this article. The final selection of sources is detailed in Table \ref{tab:challengesA}.

\section{Challenges}\label{sec:challenges}
We reviewed the literature to identify common challenges associated with integrating AI in industrial systems, as reported by multiple sources. We identified a total of 16 challenges, which we categorized into four overarching themes: system integration, data, workforce, and trustworthy AI. The subsequent section provides a detailed exploration of each identified challenge, examining their implications for the integration of AI in industrial systems.

\begin{enumerate}[label=\textbf{\arabic*}, leftmargin=0.5cm]
    \item \textbf{System Integration}
    \begin{enumerate}[label=\textbf{1.\arabic*}, leftmargin=0.26cm]
        \item \textbf{Combining OT and IT:}
        Integrating AI into technical systems presents a challenge due to the fundamental differences between Operational Technology (OT) and Information Technology (IT) \cite{Faubel.etal_2023}. 
        OT prioritizes determinism, predictability, and reliability, while IT focuses on flexibility, often through resource abstraction or containerization. 
        Furthermore, AI adds unpredictability, which makes the system less deterministic \cite{Becue.etal_2021}.
        These contrasting characteristics, coupled with the presence of outdated legacy software, make the seamless incorporation of AI into existing systems complex \cite{Borg.etal_2020, Lwakatare.etal_2019, Alenizi.etal_2023}.
        \item \textbf{AI System Infrastructure:}
        AI systems often require extensive computing resources to process large volumes of data \cite{WEF.BCG_2023, Zhou.etal_2015}, posing substantial challenges for their integration in Industry 4.0 settings \cite{Jagatheesaperumal.etal_2022, Chen.etal_2023}. Furthermore, deploying AI systems requires the use of various AI technologies and platforms, potentially leading to interoperability challenges \cite{Gabsi_2024}. Additionally, collecting data to train AI models in CPS environments usually involves updates to existing hard- and software \cite{Xu.etal_2022, Becue.etal_2021}. 
        \item \textbf{Security:} 
        The integration of data from multiple sources with emerging technologies such as AI, which is central to Industry 4.0, increases cybersecurity risks \cite{Peres.etal_2020}. Moreover, the operation of large AI models in cloud environments, rather than on-premises, exposes data flow to potential attacks \cite{Xu.etal_2022, Bajic.etal_2018}. Finally, AI systems face unique threats such as adversarial attacks and data poisoning\cite{Paleyes.etal_2022, Jagatheesaperumal.etal_2022, Becue.etal_2021}. 
        \item \textbf{Selection of AI Approaches:}
        Selecting the appropriate AI model is a common challenge \cite{Jan.etal_2023, Bajic.etal_2018,Wuest.etal_2016}. The diversity of datasets and the scarcity of CPS-specific data necessitate that AI practitioners tailor models from scratch, as no single foundational model universally excels \cite{Maschler.etal_2022}. Complementing these challenges is the process of model tuning, which involves adjusting numerous hyperparameters to optimize performance for a particular dataset, making the selection and implementation of AI approaches in CPS a complex and dynamic task \cite{He.etal_2021}.
        \item \textbf{Cost-Benefit Analysis:}
        Setting up AI systems incurs significant costs, including investments in specialized infrastructure and workforce training, which can be problematic for small and medium-sized enterprises (SMEs) \cite{Merkel-Kiss.VanGarrel_2023, Masood.Sonntag_2020, Gabsi_2024}. AI projects also demand continuous monitoring and periodic retraining of models \cite{Chen.etal_2023}, which leads to additional costs \cite{Ishikawa.Yoshioka_2019}. In safety-critical sectors, where failures carry significant costs, deriving business value from AI models can be challenging \cite{Jan.etal_2023, Becue.etal_2021}. Additionally, defining suitable key performance indicators (KPIs) in Industry 4.0 can be difficult and complicates the financial planning \cite{Machado.etal_2019}.
    \end{enumerate}
    \item \textbf{Data} 
    \begin{enumerate}[label=\textbf{2.\arabic*}, leftmargin=0.26cm]
        \item \textbf{Data Quality:}
        In practice, CPS datasets often contain noise, missing values, or incorrect timestamps due to factors such as data latency and sensor malfunctions \cite{Jan.etal_2023, Chen.etal_2023}. Furthermore, the scarcity of data on relevant events like failures leads to biased datasets  \cite{Becue.etal_2021}. Such poor data quality can directly lead to suboptimal model performance, as AI systems depend on accurate and comprehensive data to learn effectively and make precise predictions or decisions \cite{Jagatheesaperumal.etal_2022, Peres.etal_2020}.
        \item \textbf{Data Availability:}
        Many AI algorithms, especially deep learning models, depend on vast amounts of data for effective training, which can be difficult to obtain in industrial systems \cite{Maschler.etal_2022, Jagatheesaperumal.etal_2022, Chen.etal_2023}. With CPS, this data often originate from legacy systems that may not have been designed with data collection or AI integration in mind \cite{Lwakatare.etal_2019, Faubel.etal_2023, Gabsi_2024}. Publicly available datasets typically fail to serve as viable supplementary data sources. Additionally, the frequent absence of labels for supervised learning further complicates data utilization \cite{Peres.etal_2020,  Becue.etal_2021}.
        \item \textbf{Privacy:}
        Many AI models operate in the cloud rather than on local networks, necessitating the transfer of sensitive data to external servers. 
        This situation raises concerns about the unauthorized use of data without the company's consent \cite{Xu.etal_2022}. 
        Furthermore, legal obligations, such as those imposed by the General Data Protection Regulation (GDPR), further complicate data management and compliance \cite{Lavin.etal_2022, Merkel-Kiss.VanGarrel_2023}.
    \end{enumerate}
    \item \textbf{Workforce} 
    \begin{enumerate}[label=\textbf{3.\arabic*}, leftmargin=0.26cm]
        \item \textbf{Skilled Labor Shortage:}
        Implementing AI in industrial processes demands a wide range of expertise \cite{Jan.etal_2023, Becue.etal_2021}. Finding experts that are able to build, train and deploy AI models is challenging \cite{WEF.BCG_2023, Bajic.etal_2018}. Furthermore, effective data analysis demands domain-specific knowledge of the technical system \cite{Faubel.etal_2023}, further restricting the pool of qualified experts. Securing appropriate expertise is particularly challenging for SMEs \cite{Dowling.etal_2021, Merkel-Kiss.VanGarrel_2023}.
        \item \textbf{Interdisciplinary Collaboration:}
        The creation of an AI application for a technical system necessitates the collaboration of various professionals, as it involves not only AI specialists but also automation engineers, process engineers, and software engineers\cite{WEF.BCG_2023, Xu.etal_2022}.
        \item \textbf{AI Adoption:}
        The adoption of AI systems might be hindered by worker reluctance towards new technologies, which might stem from fears of job automation \cite{Gabsi_2024} and earlier unsuccessful attempts to launch AI systems \cite{WEF.BCG_2023}. This skepticism is intensified by the unpredictability of AI and occasional errors, which undermines confidence in its reliability \cite{Borg.etal_2020}. 
        Consequently, many companies remain skeptical about the near-term market advantages of implementing AI \cite{Jan.etal_2023}.
    \end{enumerate}
    \item \textbf{Trustworthy AI} 
    \begin{enumerate}[label=\textbf{4.\arabic*}, leftmargin=0.26cm]
        \item \textbf{Interpretability:}
        AI models, characterized by their intricate structures and probabilistic nature, often function as "black boxes," making it difficult to understand their decision-making processes \cite{Borg.etal_2020, Ishikawa.Yoshioka_2019}. In sectors that prioritize reliability and strict safety standards, the lack of complete transparency in AI decision-making processes can hinder its wider acceptance and integration \cite{Jagatheesaperumal.etal_2022, Perez-Cerrolaza.etal_2024, Peres.etal_2020, Gabsi_2024}.
        \item \textbf{Robustness:}
        Technical robustness is crucial for trustworthy AI \cite{AIHLEG_2019,easaEASAConceptPaper2023}. In industrial systems, a sudden decline in performance can damage machinery and endanger human safety \cite{Xu.etal_2022, Gabsi_2024}. This is especially crucial in safety-critical systems, where the lack of robustness in AI is a major concern \cite{Borg.etal_2020}. A degrading model performance can be caused by changes to the underlying data distribution, for example due to system wear or operational modifications \cite{Maschler.etal_2022, Studer.etal_2021}. This issue is compounded by the models' vulnerability to adversarial attacks \cite{Becue.etal_2021, Jagatheesaperumal.etal_2022}.
        \item \textbf{Quality Assurance:}
        Ensuring the safety of AI models and the systems they are part of is crucial, requiring comprehensive testing \cite{easaEASAConceptPaper2023}. In traditional mechanical or software engineering, the system is tested against pre-defined requirements, which is difficult to realize for AI systems \cite{Kuwajima.etal_2020}. Furthermore, AI models introduce complexities such as a lack of interpretability and inherent uncertainty in predictions, complicating the testing process \cite{Ashmore.etal_2021, Perez-Cerrolaza.etal_2024, Jan.etal_2023}. 
        Additionally, as data distributions evolve, AI models require regular retraining \cite{Xu.etal_2022} and runtime monitoring to maintain system integrity and performance \cite{Faubel.etal_2023, Ishikawa.Yoshioka_2019, Becue.etal_2021}.
        \item \textbf{Compliance:}
        The legal landscape for AI technology is still evolving, necessitating adaptability to shifting legal requirements \cite{WEF.BCG_2023}. Additionally, industrial AI systems are governed by broader regulatory frameworks in areas such as product safety and environmental impact, which can be at odds with automation via AI \cite{Becue.etal_2021}. What is missing are reliable standards to refer to \cite{vdi2022b, Faubel.etal_2023, Gabsi_2024}.
        \item \textbf{Ethical Considerations:}
        Ethical concerns in AI are growing, as highlighted by recent research and policy discussions \cite{Jan.etal_2023, AIHLEG_2019}. Adhering to ethical guidelines is essential for establishing trust among customers in AI technologies \cite{Ahlborn.etal_2019, Dowling.etal_2021}. 
        Considerations of fairness and bias are crucial in industrial AI applications as well \cite{Peres.etal_2020, Gabsi_2024}. Additionally, in scenarios critical to human safety, it may be necessary to configure AI systems explicitly to prioritize certain outcomes \cite{Perez-Cerrolaza.etal_2024}.
    \end{enumerate}
\end{enumerate}

\afterpage{
\onecolumn
\begin{ThreePartTable}
\renewcommand{\arraystretch}{1.08}
\tabcolsep1.7pt
\begin{TableNotes}[flushleft, para]
\footnotesize
\fc \: Addressed \hc \:  mentioned \ec \: not covered. 
\end{TableNotes}
\centering
\footnotesize
\begin{xltabular}{\linewidth}{lc|ccccc|ccc|ccc|ccccc|}
\caption{A table summarizing AI integration challenges from the literature.}\label{tab:challengesA}\\
       & 
       & \multicolumn{5}{c}{\textbf{System Integration}} \vline 
       & \multicolumn{3}{c}{\textbf{Data}} \vline 
       & \multicolumn{3}{c}{\textbf{Workforce}} \vline
       & \multicolumn{5}{c}{\textbf{Trustworthy AI}} \vline\\ 
       \cmidrule{3-18}
       & 
       &\rotatebox[origin=l]{90}{Combining OT and IT}  
       &\rotatebox[origin=l]{90}{AI System Infrastructure}  
       &\rotatebox[origin=l]{90}{Security} 
       &\rotatebox[origin=l]{90}{Selection of AI Approaches} 
       &\rotatebox[origin=l]{90}{Cost-Benefit Analysis} 
       &\rotatebox[origin=l]{90}{Data Quality} 
       &\rotatebox[origin=l]{90}{Data Availability} 
       &\rotatebox[origin=l]{90}{Privacy}
       &\rotatebox[origin=l]{90}{Skilled Labor Shortage} 
       &\rotatebox[origin=l]{90}{Interdisciplinary Collaboration} 
       &\rotatebox[origin=l]{90}{AI Adoption}
       &\rotatebox[origin=l]{90}{Interpretability}  
       &\rotatebox[origin=l]{90}{Robustness}  
       &\rotatebox[origin=l]{90}{Quality Assurance} 
       &\rotatebox[origin=l]{90}{Compliance}
       &\rotatebox[origin=l]{90}{Ethical Considerations}
       \\
       \midrule
       & \multicolumn{1}{c|}{\diagbox[width=10em]{\textbf{References}}{\textbf{Challenges}}}
       & \textbf{1.1} & \textbf{1.2} & \textbf{1.3} & \textbf{1.4} & \textbf{1.5} 
       & \textbf{2.1} & \textbf{2.2} & \textbf{2.3}  
       & \textbf{3.1} & \textbf{3.2} & \textbf{3.3} 
       & \textbf{4.1} & \textbf{4.2} & \textbf{4.3} & \textbf{4.4} & \textbf{4.5}\\ 
       \midrule
\endfirsthead

\caption[]{(continued from previous page)} \\
       & 
       & \multicolumn{5}{c}{\textbf{System Integration}} \vline 
       & \multicolumn{3}{c}{\textbf{Data}} \vline 
       & \multicolumn{3}{c}{\textbf{Workforce}} \vline
       & \multicolumn{5}{c}{\textbf{Trustworthy AI}} \vline\\ 
       \cmidrule{3-18}
       & 
       &\rotatebox[origin=l]{90}{Combining OT and IT}  
       &\rotatebox[origin=l]{90}{AI System Infrastructure}  
       &\rotatebox[origin=l]{90}{Security} 
       &\rotatebox[origin=l]{90}{Selection of AI Approaches} 
       &\rotatebox[origin=l]{90}{Cost-Benefit Analysis} 
       &\rotatebox[origin=l]{90}{Data Quality} 
       &\rotatebox[origin=l]{90}{Data Availability} 
       &\rotatebox[origin=l]{90}{Privacy}
       &\rotatebox[origin=l]{90}{Skilled Labor Shortage} 
       &\rotatebox[origin=l]{90}{Interdisciplinary Collaboration} 
       &\rotatebox[origin=l]{90}{AI Adoption}
       &\rotatebox[origin=l]{90}{Interpretability}  
       &\rotatebox[origin=l]{90}{Robustness}  
       &\rotatebox[origin=l]{90}{Quality Assurance} 
       &\rotatebox[origin=l]{90}{Compliance}
       &\rotatebox[origin=l]{90}{Ethical Considerations}
       \\
       \midrule
       & \multicolumn{1}{c|}{\diagbox[width=10em]{\textbf{References}}{\textbf{Challenges}}}
       & \textbf{1.1} & \textbf{1.2} & \textbf{1.3} & \textbf{1.4} & \textbf{1.5} 
       & \textbf{2.1} & \textbf{2.2} & \textbf{2.3}  
       & \textbf{3.1} & \textbf{3.2} & \textbf{3.3} 
       & \textbf{4.1} & \textbf{4.2} & \textbf{4.3} & \textbf{4.4} & \textbf{4.5}\\ 
       \midrule
\endhead

\bottomrule
\multicolumn{18}{r}{\footnotesize \fc \: Addressed \hc \:  mentioned \ec \: not covered. \hfill to be continued on the next page}
\endfoot
\bottomrule
\insertTableNotes
\endlastfoot
        & \multicolumn{1}{l|}{Ahlborn et al. (2019) \cite{Ahlborn.etal_2019}} & \ec & \hc & \hc & \ec & \ec & \fc & \fc & \fc & \hc & \hc & \ec & \fc & \fc & \hc & \hc & \fc \\ 
        & \multicolumn{1}{l|}{AI HLEG (2019) \cite{AIHLEG_2019}} & \ec & \ec & \hc & \ec & \ec & \fc & \ec & \hc & \ec & \ec & \fc & \fc & \fc & \fc & \hc & \fc \\ 
        & \multicolumn{1}{l|}{Alenizi et al. (2019) \cite{Alenizi.etal_2023}} & \fc & \hc & \fc & \ec & \fc & \fc & \fc & \hc & \hc & \hc & \hc & \ec & \fc & \fc & \hc & \hc \\ 
        & \multicolumn{1}{l|}{Amershi et al. (2019) \cite{Amershi.etal_2019}} & \hc & \fc & \ec & \hc & \ec & \fc & \fc & \hc & \ec & \fc & \ec & \fc & \hc & \ec & \fc & \ec \\ 
        & \multicolumn{1}{l|}{Arinez et al. (2019) \cite{Arinez.etal_2020}} & \hc & \hc & \hc & \fc & \ec & \fc & \fc & \ec & \ec & \hc & \ec & \fc & \ec & \ec & \ec & \ec \\ 
        & \multicolumn{1}{l|}{Arnold et al. (2019) \cite{Arnold.etal_2019}} & \ec & \ec & \fc & \ec & \ec & \hc & \hc & \ec & \ec & \ec & \fc & \fc & \fc & \fc & \hc & \hc \\ 
        & \multicolumn{1}{l|}{Ashmore et al. (2021) \cite{Ashmore.etal_2021}} & \ec & \ec & \fc & \fc & \ec & \fc & \hc & \ec & \ec & \ec & \ec & \fc & \fc & \fc & \ec & \hc \\ 
        & \multicolumn{1}{l|}{Bajic et al. (2018) \cite{Bajic.etal_2018}} & \ec & \hc & \fc & \fc & \ec & \fc & \fc & \fc & \fc & \ec & \ec & \ec & \ec & \ec & \ec & \ec \\ 
        & \multicolumn{1}{l|}{B{\'e}cue et al. (2021) \cite{Becue.etal_2021}} & \fc & \fc & \fc & \hc & \fc & \fc & \fc & \hc & \fc & \ec & \ec & \ec & \hc & \fc & \fc & \ec \\ 
        & \multicolumn{1}{l|}{Belani et al. (2019) \cite{Belani.etal_2019}} & \ec & \fc & \fc & \ec & \fc & \ec & \fc & \fc & \fc & \fc & \ec & \fc & \ec & \fc & \fc & \hc \\ 
        & \multicolumn{1}{l|}{Bertolini et al. (2021) \cite{Bertolini.etal_2021}} & \ec & \ec & \ec & \fc & \ec & \hc & \hc & \ec & \hc & \ec & \ec & \hc & \ec & \ec & \ec & \ec \\ 
        & \multicolumn{1}{l|}{Borg et al. (2020) \cite{Borg.etal_2020}} & \fc & \ec & \hc & \hc & \ec & \hc & \fc & \ec & \ec & \ec & \hc & \fc & \fc & \fc & \fc & \ec \\
        & \multicolumn{1}{l|}{Bosch et al. (2020) \cite{Bosch.etal_2020}} & \hc & \fc & \ec & \hc & \ec & \fc & \fc & \hc & \ec & \hc & \ec & \hc & \hc & \hc & \ec & \ec \\
        & \multicolumn{1}{l|}{Chen et al. (2023) \cite{Chen.etal_2023}} & \ec & \fc & \hc & \hc & \hc & \fc & \fc & \hc & \ec & \ec & \ec & \fc & \hc & \hc & \ec & \ec \\
        & \multicolumn{1}{l|}{DIN (2019) \cite{DINSPEC9200112019}} & \ec & \ec & \hc & \fc & \ec & \fc & \fc & \hc & \ec & \ec & \hc & \fc & \fc & \fc & \hc & \hc \\ 
        & \multicolumn{1}{l|}{DIN and DKE (2022) \cite{vdi2022b}} & \fc & \ec & \fc & \ec & \ec & \fc & \fc & \hc & \hc & \hc & \hc & \fc & \fc & \fc & \hc & \fc \\ 
        & \multicolumn{1}{l|}{Dowling et al (2021) \cite{Dowling.etal_2021}} & \ec & \ec & \fc & \ec & \fc & \fc & \fc & \fc & \fc & \fc & \fc & \hc & \hc & \ec & \fc & \fc \\ 
        & \multicolumn{1}{l|}{EASA (2023) 
        \cite{easaEASAConceptPaper2023}} & \ec & \ec & \hc & \ec & \ec & \fc & \ec & \hc & \ec & \ec & \fc & \fc & \fc & \fc & \fc & \hc \\ 
        & \multicolumn{1}{l|}{Ebers (2022) \cite{Ebers_2022}} & \ec & \fc & \ec & \ec & \ec & \ec & \ec & \fc & \ec & \ec & \ec & \fc & \ec & \fc & \fc & \fc \\ 
        & \multicolumn{1}{l|}{Faubel et al. (2023) \cite{Faubel.etal_2023}} & \fc & \fc & \ec & \ec & \hc & \hc & \fc & \hc & \hc & \ec & \ec & \ec & \hc & \fc & \hc & \ec \\ 
        & \multicolumn{1}{l|}{Gabsi (2024) \cite{Gabsi_2024}} & \hc & \fc & \hc & \ec & \fc & \hc & \fc & \hc & \fc & \hc & \fc & \fc & \fc & \hc & \fc & \fc \\ 
        & \multicolumn{1}{l|}{Gebru et al. (2021) \cite{Gebru.etal_2021}} & \ec & \ec & \fc & \ec & \ec & \fc & \fc & \hc & \ec & \ec & \fc & \fc & \fc & \fc & \ec & \fc \\ 
        & \multicolumn{1}{l|}{He et al. (2021) \cite{He.etal_2021}} & \ec & \ec & \ec & \ec & \fc & \fc & \fc & \ec & \fc & \ec & \ec & \hc & \fc & \ec & \ec & \ec \\ 
        & \multicolumn{1}{l|}{IEC 2018 \cite{IEC_2018}} & \hc & \fc & \fc & \ec & \ec & \fc & \fc & \fc & \ec & \hc & \ec & \fc & \fc & \fc & \fc & \fc \\ 
        & \multicolumn{1}{l|}{ISO/IEC (2022) \cite{isoiec2022a}} & \hc & \ec & \fc & \ec & \ec & \fc & \hc & \fc & \hc & \hc & \ec & \hc & \fc & \hc & \fc & \fc \\ 
        & \multicolumn{1}{l|}{ISO/IEC (2024) \cite{isoiec2024}} & \fc & \fc & \hc & \fc & \ec & \fc & \fc & \fc & \ec & \hc & \ec & \fc & \fc & \fc & \fc & \hc \\ 
        & \multicolumn{1}{l|}{Ishikawa and Yoshioka (2019) \cite{Ishikawa.Yoshioka_2019}} & \ec & \ec & \hc & \hc & \fc & \ec & \hc & \hc & \ec & \hc & \ec & \fc & \hc & \fc & \ec & \ec \\ 
        & \multicolumn{1}{l|}{Jagatheesaperum et al. (2022) \cite{Jagatheesaperumal.etal_2022}} & \hc & \fc & \fc & \ec & \ec & \fc & \fc & \fc & \ec & \ec & \ec & \fc & \hc & \hc & \fc & \ec \\ 
        & \multicolumn{1}{l|}{Jan et al. (2023) \cite{Jan.etal_2023}} & \ec & \ec & \fc & \fc & \fc & \fc & \fc & \fc & \fc & \ec & \fc & \fc & \ec & \fc & \ec & \fc \\
        & \multicolumn{1}{l|}{Kuwajima et al. (2020) \cite{Kuwajima.etal_2020}} & \ec & \ec & \ec & \ec & \ec & \hc & \ec & \ec & \ec & \ec & \ec & \fc & \fc & \fc & \hc & \ec \\ 
        & \multicolumn{1}{l|}{Lavin et al. (2022) \cite{Lavin.etal_2022}} & \fc & \hc & \fc & \ec & \fc & \fc & \fc & \fc & \ec & \fc & \fc & \fc & \fc & \hc & \fc & \fc \\ 
        & \multicolumn{1}{l|}{Lee et al. (2018) \cite{Lee.etal_2018}} & \hc & \ec & \fc & \ec & \ec & \fc & \hc & \ec & \hc & \ec & \ec & \ec & \fc & \ec & \ec & \ec \\ 
        & \multicolumn{1}{l|}{Lwakatare et al. (2019) \cite{Lwakatare.etal_2019}} & \hc & \fc & \ec & \fc & \ec & \fc & \fc & \fc & \ec & \ec & \ec & \hc & \fc & \fc & \hc & \ec \\ 
        & \multicolumn{1}{l|}{Machado et al. (2019) \cite{Machado.etal_2019}} & \ec & \hc & \hc & \hc & \fc & \ec & \ec & \hc & \fc & \fc & \fc & \ec & \ec & \ec & \fc & \ec \\
        & \multicolumn{1}{l|}{Maschler et al. (2022) \cite{Maschler.etal_2022}} & \ec & \ec & \ec & \ec & \hc & \hc & \fc & \fc & \ec & \ec & \ec & \ec & \fc & \ec & \ec & \ec \\ 
        & \multicolumn{1}{l|}{Masood and Sonntag (2020) \cite{Masood.Sonntag_2020}} & \ec & \hc & \hc & \hc & \fc & \ec & \ec & \ec & \fc & \ec & \hc & \ec & \ec & \ec & \ec & \ec \\
        & \multicolumn{1}{l|}{Merkel-Kiss and Von Garrel (2023) \cite{Merkel-Kiss.VanGarrel_2023}} & \hc & \hc & \hc & \hc & \fc & \hc & \hc & \hc & \fc & \ec & \hc & \hc & \hc & \hc & \hc & \ec \\ 
        & \multicolumn{1}{l|}{Metternich et al. (2021) \cite{Metternich.etal_2021}} & \ec & \fc & \ec & \hc & \fc & \fc & \fc & \hc & \fc & \fc & \hc & \hc & \hc & \ec & \fc & \ec \\ 
        & \multicolumn{1}{l|}{Mitchell et al. (2019) \cite{Mitchell.etal_2019}} & \ec & \ec & \ec & \ec & \ec & \hc & \hc & \hc & \ec & \ec & \hc & \hc & \fc & \fc & \fc & \fc \\ 
        & \multicolumn{1}{l|}{Niggemann et al. (2023) \cite{Niggemann.etal_2023}} & \ec & \ec & \ec & \ec & \ec & \fc & \fc & \ec & \ec & \fc & \ec & \fc & \ec & \hc & \ec & \ec \\ 
        & \multicolumn{1}{l|}{Nikolic et al. (2017) \cite{Nikolic.etal_2017}} & \hc & \ec & \fc & \hc & \hc & \hc & \fc & \ec & \hc & \ec & \ec & \ec & \hc & \ec & \ec & \ec \\ 
        & \multicolumn{1}{l|}{Nti et al. (2022) \cite{Nti.etal_2022}} & \ec & \ec & \ec & \fc & \hc & \ec & \ec & \ec & \ec & \fc & \ec & \ec & \ec & \fc & \ec & \ec \\ 
        & \multicolumn{1}{l|}{OECD (2019) \cite{oecd2019a}} & \ec & \ec & \fc & \ec & \ec & \ec & \ec & \fc & \ec & \ec & \ec & \fc & \hc & \ec & \fc & \fc \\ 
        & \multicolumn{1}{l|}{Paleyes et al. (2022) \cite{Paleyes.etal_2022}} & \ec & \hc & \fc & \fc & \fc & \fc & \fc & \fc & \fc & \fc & \fc & \fc & \hc & \fc & \fc & \fc \\ 
        & \multicolumn{1}{l|}{Pereira and Thomas (2020) \cite{Pereira.Thomas_2020}} & \ec & \fc & \fc & \fc & \ec & \fc & \fc & \ec & \ec & \ec & \ec & \fc & \fc & \fc & \ec & \ec \\ 
        & \multicolumn{1}{l|}{Peres et al. (2020) \cite{Peres.etal_2020}} & \ec & \ec & \fc & \ec & \ec & \fc & \fc & \fc & \ec & \ec & \fc & \fc & \ec & \hc & \ec & \fc \\ 
        & \multicolumn{1}{l|}{Perez-Cerrolaz et al. (2024) \cite{Perez-Cerrolaza.etal_2024}} & \fc & \hc & \hc & \hc & \ec & \fc & \hc & \ec & \ec & \ec & \ec & \fc & \fc & \fc & \fc & \fc \\ 
        & \multicolumn{1}{l|}{Ruf et al. (2021) \cite{Ruf.etal_2021}} & \ec & \ec & \ec & \ec & \ec & \fc & \ec & \hc & \ec & \fc & \ec & \ec & \fc & \fc & \ec & \ec \\ 
        & \multicolumn{1}{l|}{Sculley et al. (2015) \cite{Sculley.etal_2015}} & \ec & \hc & \ec & \ec & \fc & \hc & \hc & \ec & \ec & \ec & \ec & \ec & \fc & \hc & \ec & \ec \\ 
        & \multicolumn{1}{l|}{Studer et al. (2021) \cite{Studer.etal_2021}} & \ec & \ec & \fc & \ec & \ec & \fc & \fc & \ec & \ec & \ec & \fc & \hc & \fc & \fc & \fc & \hc \\ 
        & \multicolumn{1}{l|}{Treveil et al. (2020) \cite{Treveil.etal_2020}} & \ec & \fc & \fc & \ec & \ec & \fc & \fc & \hc & \ec & \fc & \fc & \fc & \fc & \hc & \fc & \hc \\ 
        & \multicolumn{1}{l|}{WEF and BCG (2023) \cite{WEF.BCG_2023}} & \hc & \fc & \fc & \hc & \ec & \hc & \hc & \ec & \fc & \fc & \fc & \ec & \ec & \ec & \fc & \ec \\     
        & \multicolumn{1}{l|}{Wuest et al. (2016) \cite{Wuest.etal_2016}} & \ec & \hc & \hc & \fc & \ec & \fc & \fc & \ec & \ec & \ec & \ec & \fc & \hc & \ec & \ec & \ec \\     
        & \multicolumn{1}{l|}{Xu et al. (2022) \cite{Xu.etal_2022}} & \ec & \fc & \fc & \hc & \ec & \fc & \fc & \fc & \ec & \fc & \ec & \ec & \fc & \ec & \ec & \ec \\ 
        & \multicolumn{1}{l|}{Zhou et al. (2015) \cite{Zhou.etal_2015}} & \fc & \hc & \hc & \ec & \ec & \ec & \ec & \hc & \ec & \ec & \ec & \ec & \ec & \fc & \hc & \ec \\
\end{xltabular}
\end{ThreePartTable}
}

Table \ref{tab:challengesA} summarizes the challenges and gives an overview about the relevant literature that addresses or mentions them. In this context, 'addressed' indicates that the source explicitly focuses on the specified challenge, whereas 'mentioned' denotes sources that refer to the challenge without emphasizing it or recognize closely related issues.

\section{Discussion}\label{sec:discussion}

\begin{figure}[!htb]
\centering
\includegraphics[scale=.6]{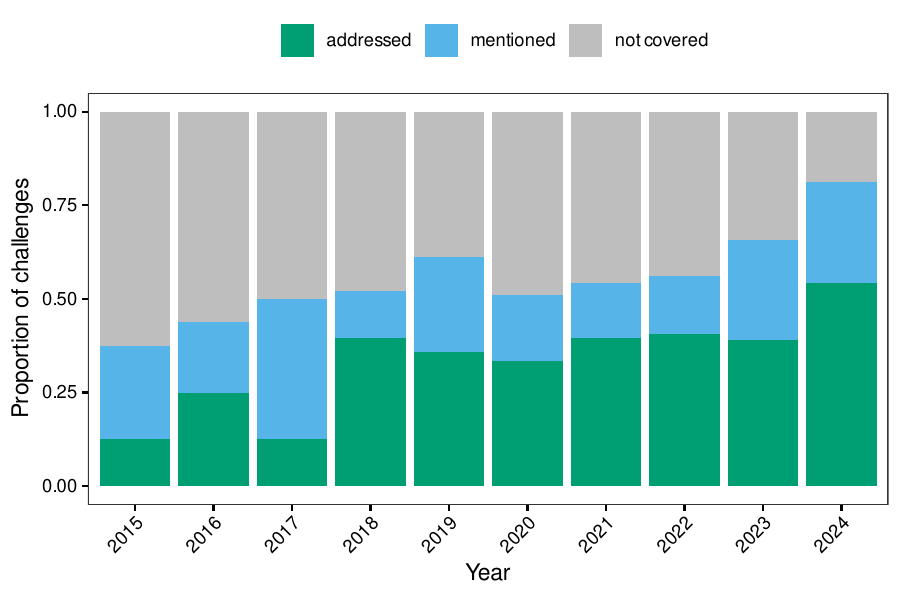}
\caption{Proportion of addressed, mentioned, and not covered challenges over time (sample size $n=55$).
'Addressed' means explicit focus on the challenge, while 'mentioned' indicates brief reference to it or related issues.}
\label{fig:proportion}
\end{figure}

\begin{figure}[!htb]
\centering
\includegraphics[scale=.65]{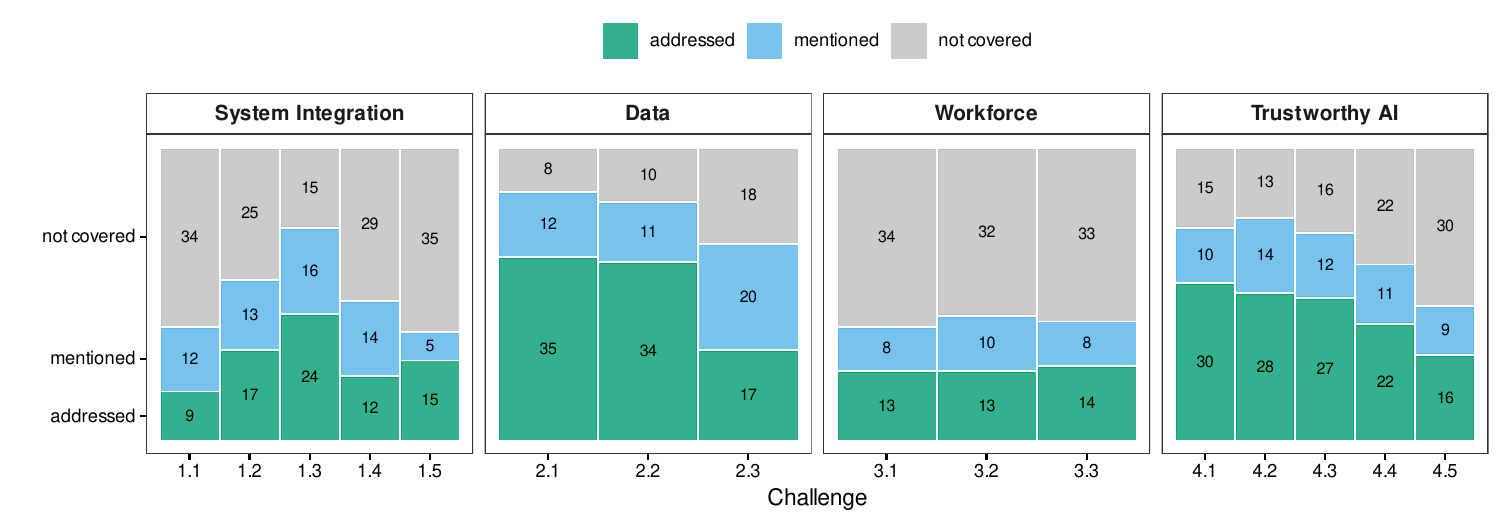}
\caption{Mosaic plot of the identified challenges presented in Table \ref{tab:challengesA} 
(sample size $n=55$).
'Addressed' means explicit focus on the challenge, while 'mentioned' indicates brief reference to it or related issues.}
\label{fig:mosaic}
\end{figure}

This survey reviewed recent literature to identify the current challenges relevant to contemporary industrial applications. As AI technology evolves, new challenges have emerged that have not been recognized by earlier sources, as illustrated in Figure \ref{fig:proportion}. The increasing proportion of literature addressing and mentioning the identified challenges suggests a growing academic interest in these challenges, with earlier years featuring a more limited number of studies, as Figure \ref{fig:num_of_papers} demonstrates. Notably, the frequency with which various challenges are mentioned and addressed in the literature varies considerably, as shown in Figure \ref{fig:mosaic}. The most frequently cited challenges primarily fall into two main categories: data and trustworthy AI. 

Data is fundamental to the development and performance of AI systems. In the reviewed literature, data quality and data availability are among the most often addressed or mentioned challenges. Many CPS datasets are derived from legacy systems not initially intended for AI data collection \cite{Lwakatare.etal_2019, Faubel.etal_2023, Gabsi_2024}. Data latency issues and sensor malfunctions then degrade the data quality \cite{Jan.etal_2023, Chen.etal_2023}. Addressing these challenges often requires extensive data preprocessing or significant hardware upgrades. Various strategies have been proposed to mitigate these issues, including synthetic data generation through simulations \cite{Zhang.etal_2022, Krantz.etal_2022} and federated learning \cite{Li.etal_2020, Savazzi.etal_2021}, which allows for the utilization of data across decentralized systems while managing privacy concerns.

Trustworthy AI is increasingly important as AI becomes integrated into more systems. The most addressed or mentioned challenges in this area are robustness, interpretability and quality assurance. These issues are especially critical in Industry 4.0, due to stringent regulatory environments governing technical systems. New regulatory frameworks for AI, such as the AI Act in the European Union \cite{EC_2024} and similar proposals in the United States \cite{WH_2022} and the United Kingdom \cite{UK_2023}, underscore the urgency of these concerns. The lack of robustness of AI systems makes a thorough quality assurance crucial, yet the lack of interpretability and the dependency on data complicate testing efforts. Approaches such as advanced testing frameworks \cite{Dix.etal_2023, Windmann.etal_2023}, explainable AI (XAI) \cite{Ahmed.etal_2022, Trivedi.etal_2024}, and the development of new process models \cite{Lavin.etal_2022, Mitchell.etal_2019} aim to improve AI system robustness, clarify decision-making processes, and meet regulatory demands, thereby enhancing overall system quality and accountability.

The reviewed literature less frequently addressed system integration and workforce challenges.
However, practitioners, especially in SMEs, often highlight challenges such as financial constraints, selecting the appropriate AI approach, and addressing the skilled labor shortage \cite{Merkel-Kiss.VanGarrel_2023, Masood.Sonntag_2020}. This discrepancy suggests a gap between academic research and practical industry needs. Addressing these barriers through approaches such as automated machine learning (AutoML) \cite{He.etal_2021, Zoeller.Huber_2021}, transfer learning \cite{Li.etal_2022, Maschler.etal_2022}, and enhanced standardization \cite{vdi2022b} is thus essential for enabling the successful adoption of AI technologies in the industrial sector. In order to design a holistic solution approach, a comprehensive understanding of all potential challenges is necessary.

While our review highlights key challenges of integrating AI in industrial systems, it is crucial to acknowledge the limitations of our study. Our review was dependent on published literature, standards, and reports, which may not fully capture the rapidly evolving landscape of AI technologies and their applications in industry. Furthermore, existing research may underrepresent emerging challenges. Recognizing these gaps, future research should incorporate a broader spectrum of sources, including more recent case studies and primary data, to provide a more comprehensive understanding of the challenges and opportunities in AI integration. This will not only enhance the credibility of our analysis but also ensure that the discussion remains relevant and informative for guiding both practitioners and researchers in addressing these challenges.

\section{Conclusion}\label{sec:conclusion}

This paper identified and categorized key challenges associated with integrating AI in Industry 4.0, which might explain the underwhelming impact AI has had on the sector so far. Each challenge was linked to relevant sources, allowing readers interested in specific issues to explore them further. 
Through quantitative analysis, it highlights critical areas such as data-related challenges and the development of trustworthy AI. Promising research directions include improving the availability of high-quality industrial data for AI models and enhancing quality assurance techniques to ensure the robustness of AI models. Additionally, the establishment of clearer technical standards is essential to prepare for increasingly stringent regulations.

Furthermore, the study identifies substantial research gaps, especially in system integration and workforce-related issues. These areas, though crucial to practitioners, have received less attention by the reviewed literature. SMEs, in particular, would benefit from more accessible and efficient tools to integrate AI systems with existing legacy systems. Techniques like AutoML and transfer learning promise increased accessibility, thereby mitigating the impact of the prevailing talent shortage.

In summary, this paper serves as a resource for the AI and industrial automation community, clarifying and prioritizing challenges for further research and practical application. By addressing these challenges, the field can advance towards more effective and successful integration of AI technologies in Industry 4.0.

\section*{Acknowledgment}

This research as part of the projects SmartShip and EKI is funded by dtec.bw – Digitalization and Technology Research Center of the Bundeswehr which we gratefully acknowledge. dtec.bw is funded by the European Union – NextGenerationEU.

\bibliographystyle{myIEEEtran}

\end{document}